\documentclass[a4paper, 10pt, conference]{ieeeconf}      

\IEEEoverridecommandlockouts                              

\usepackage{amsfonts}
\usepackage{algorithm}  
\usepackage{algorithmic}
\usepackage{graphicx}
\usepackage{subfigure}
\usepackage{makecell}

\title{\LARGE \bf
ACDER: Augmented Curiosity-Driven Experience Replay
}

\author{Boyao Li$^{1,2}$, Tao Lu$^{2}$, Jiayi Li$^{1,2}$, Ning Lu$^{1,2}$, Yinghao Cai$^{2}$, and Shuo Wang$^{1,2}$
\thanks{*This work was supported by the National Key R\emph{\&}D Program of China (grant 2019YFB1311901) and partly by Science and Technology on Space Intelligent Control Laboratory for National Defense, No. KGJZDSYS-2018-09, the National Natural Science Foundation of China under Grant U1713222, 61773378}
\thanks{$^{1}$School of Artificial Intelligence, University of Chinese Academy of Sciences, Beijing, China}
\thanks{$^{2}$Research Center on Intelligent Robotic Systems, Institute of Automation, Chinese Academy of Sciences, Beijing, China
        {\tt\small \{liboyao2017, tao.lu\}@ia.ac.cn}}%
}
\usepackage[letterpaper,top=72pt,bottom=54pt,left=54pt,right=54pt]{geometry}
\begin{document}
\maketitle
\thispagestyle{empty}
\pagestyle{empty}

\begin{abstract}
Exploration in environments with sparse feedback remains a challenging research problem in reinforcement learning (RL). When the RL agent explores the environment randomly, it results in low exploration efficiency, especially in robotic manipulation tasks with high dimensional continuous state and action space. In this paper, we propose a novel method, called Augmented Curiosity-Driven Experience Replay (ACDER), which leverages (i) a new goal-oriented curiosity-driven exploration to encourage the agent to pursue novel and task-relevant states more purposefully and (ii) the dynamic initial states selection as an automatic exploratory curriculum to further improve the sample-efficiency. Our approach complements Hindsight Experience Replay (HER) by introducing a new way to pursue valuable states. Experiments conducted on four challenging robotic manipulation tasks with binary rewards, including Reach, Push, Pick$\emph{\&}$Place and Multi-step Push. The empirical results show that our proposed method significantly outperforms existing methods in the first three basic tasks and also achieves satisfactory performance in multi-step robotic task learning.

\end{abstract}
\section{INTRODUCTION}
Deep reinforcement learning (DRL) has achieved remarkable results in learning policies for sequential decision-making problems. This includes the domains ranging from playing Atari games [1] to high dimensional continuous control [2, 3] and learning robotic manipulation tasks [4, 5, 6, 7].

Despite these successes, a common challenge, especially in robotic tasks, is to make the agent learn sample-efficiently in environments with sparse rewards [8]. Due to receiving bonuses only when completing tasks, few valuable experiences contribute to guiding policy optimization [9], which results in high sample complexity. With HER [10], the agent has the capability to transform the unsuccessful experiences into successful ones and gets more valuable samples to encourage learning. However, this method still suffers from low sample efficiency, mainly because of random exploration [11, 12]. With increasing task complexity and state-space dimensionality, it is difficult to visit the whole environment by taking random sequences of actions in robotic manipulation tasks [13]. Therefore, developing a method that can guide exploration in a certain directed way is necessary to urge the agent to pursue states unseen before rather than repeating those visited.

One ability we humans have, unlike the standard RL agent, is to learn with curiosity [14]. Imagine that you are looking for a watch in an unknown room. Instead of seeking at will, you always first choose an area where you think is most likely to find and then look for it nearby. If fail, you will try it again in another new place until finally succeed. In this way, you will look all over the room faster to achieve the goal and not waste time in the place searched before. Inspired by this, the kind of curiosity-driven exploration [15] might make the exploratory process more efficient and impel the agent to pursue those unseen states. One recent and popular method, intrinsic curiosity module (ICM) [16] augments task reward with intrinsic curiosity-driven reward based on state prediction error to achieve this objective. However, it must be noted that blindly pursuing novel states also makes sample inefficiency and suffers from the “couch-potato” issues, which hinders the agent from learning [17].

In this paper, we introduce a method called Augmented Curiosity-Driven Experience Replay (ACDER). This technique aims to guide the agent to explore the environment more efficiently and purposefully. First, we propose a new goal-oriented curiosity-driven exploration method, in which we employ a simplified intrinsic curiosity module to generate intrinsic curiosity reward for each training transition based on its state prediction error and introduce a goal-oriented factor to guarantee the sampled transitions modified by HER to be relevant to the tasks. Combining curiosity rewards and the new goal-oriented task rewards, the agent prefers to seek task-relevant and unseen states before more often under ensuring success. Second, instead of always producing episodes from a fixed state, we adopt a dynamic initial state selection mechanism to start the agent from novel states with high curiosity rewards. We can see this as an automatic exploratory curriculum which gradually explores the environment in a directed way according to the agent's current knowledge about it and even visit those beyond the originally reachable states. We evaluate ACDER with Deep Deterministic Policy Gradient (DDPG) [18] on the robotic manipulation tasks. Ablation studies show that each component of our method plays a positive role in achieving a higher success rate and faster convergence. Moreover, the policies trained in simulation with our method are deployed on a physical robot and can complete the tasks successfully.

\section{RELATED WORK}

\subsection{Exploration}
In reinforcement learning, the basic idea of exploration is to encourage the agent to explore novel or uncertain states rarely visited before to gain more information about the environment and accelerate policy learning. Current exploration methods can be classified into count-based exploration and curiosity-driven exploration. Count-based exploration methods aim to utilize the visitation counts as an intrinsic bonus to instruct the agent to explore the environment. The classic techniques directly turn the number of times states [19] or state-action pairs [20, 21] visited into bonuses, especially for discrete observation space. Extended to high-dimensional continuous state space, there is less chance of appearing two identical states in the environment, so other methods are proposed, such as fitting a density generative model to estimate pesudo-count [22, 23], mapping states to hash codes to count their occurrences with a hash table [24] or discriminating each visited state against all others implicitly based on an exemplar model [25]. Curiosity-driven exploration [26, 27] formulates state novelty as the intrinsic reward. In [16], ICM rewards the agent based on the state prediction error [28, 29]. When visiting unseen states, the prediction error becomes high, and the agent receives high intrinsic rewards. And in [17], it compares current observation with those stored in episodic memory to generate bonuses to solve the “couch-potato” problems. However, most current curiosity-driven methods are restricted to video game environments and combined with on-policy RL algorithms.

\subsection{Experience Replay}
The technique of experience replay has been introduced in [30] and becomes to be applied due to the success of DQN [1] in playing Atari. It focuses on selecting which experiences to store and how to use them to accelerate the training procedure [31, 32, 33]. Many studies have improved the experience replay. Prioritized Experience Replay [34, 35] prioritizes the transitions with high TD-error in the replay buffer more frequently to speed up learning. HER [10] replaces original goals with achieved goals to encourage the agent to learn much from the undesired outcome. Based on HER, Dynamic Hindsight Experience Replay [36] is proposed to assemble successful experiences from two relevant failure to deal with robotic tasks with dynamic goals; Energy-based Prioritization method [37] prioritizes trajectories in which more physical work-energy is done and Curiosity-driven Prioritization method [38] prioritizes trajectories in which rare goal states are achieved. Currently, most improvements of experience replay are made to solve how to use transitions in the replay buffer, not how to generate episodes to store.

\section{BACKGROUND}

In this section, we provide an introduction to the preliminaries, including the basic concepts for reinforcement learning, Deep Deterministic Policy Gradient, and Hindsight Experience Replay.

\subsection{Reinforcement Learning}
We consider the standard Markov Decision Process (MDP) represented as a tuple $ (\emph{\textbf{S}}, \emph{\textbf{A}}, \emph{\textbf{P}}, \emph{\textbf{R}}, \gamma)$, which consists of a set of states $\emph{\textbf{S}}$, a set of actions $\emph{\textbf{A}}$, a transition function $\emph{\textbf{P}}(s'|s,a)$, a reward function $\emph{\textbf{R}}(s,a)$ and a discount factor $\gamma$. An agent interacts with environment $\emph{\textbf{E}}$ which is assumed to be fully observable. At each timestep $\emph{t}$, the agent performs an action $\emph{a}_t \in \emph{\textbf{A}}$ at current state $\emph{s}_t\in\emph{\textbf{S}}$ and receives a reward $\emph{r}_t\in\emph{\textbf{R}}$ provided by the environment and the next state $\emph{s}_{t+1}$ sampled from the distribution $\emph{\textbf{P}}(s_{t+1}|\emph{s}_t, \emph{a}_t)$. The agent chooses actions based on a policy $\emph{a}_t=\pi(\emph{s}_t)$, which maps from states to actions directly. The goal in reinforcement learning is to learn an optimal policy $\pi$ to maximize the accumulated reward, i.e. the expected return $\emph{R}_t=\sum_{i=t}^T\gamma^{i-t}\emph{r}_i$, over time horizon $\emph{T}$ and with the discount factor $\gamma \in[0,1]$.

\subsection{Deep Deterministic Policy Gradient}
Deep Deterministic Policy Gradient (DDPG), an off-policy model-free reinforcement learning algorithm, has shown promising performance in continuous control tasks. DDPG utilizes an actor-critic framework, which incorporates policy gradient methods with value approximation methods for RL: an actor network approximating the policy $\pi:\mathcal{S}\rightarrow\mathcal{A}$ and a critic network learning the action-value function $Q:\mathcal{S}\times\mathcal{A}\rightarrow\mathbb{R}$. The critic is trained using temporal difference method to minimize the loss $\emph{L}$ w.r.t $\theta^Q$: 
$$
\emph{L}=\frac{1}{N}\sum_{i}(y_{i}-Q(s_{i},a_{i}|\theta^Q))^2 \eqno{(1)}
$$
where
$$
y_{i}=r_{i}+\gamma Q'(s_{i+1},\pi'(s_{i+1}|\theta^{\pi'})|\theta^{Q'}) \eqno{(2)}
$$
The actor is trained using policy gradient to maximize $Q^\pi$ w.r.t $\theta^\pi$:
$$
\nabla_{\theta^\pi}\emph{J}\approx\mathbb{E}_{s_{t}\sim E}[\nabla_{\theta^\pi}Q(s,a|\theta^Q)|_{s=s_{t},a=\pi(s_t|\theta^\pi)}] \eqno{(3)}
$$
Considering stability reasons, separate slower moving target networks $\pi'$ with parameters $Q^{\pi'}$ and $Q'$ are used to compute the target $y_{i}$. DDPG maintains a replay buffer to store transitions $(s_{t},a_{t},s_{t+1},r_{t})$ experienced before to allow the algorithm updating from a set of uncorrected transitions, which stabilizes training by breaking the temporal correlations of the update.

\subsection{Hindsight Experience Replay}
Hindsight Experience Replay (HER) is presented to solve sparse-reward challenge in reinforcement learning and can be used with any off-policy RL algorithm. HER is based on training universal policies [39] which take as input not only the current state but also a goal state. The pivotal idea is: after experiencing some episodes, every transition $(s_{t},a_{t},g_{t},s_{t+1},r_{t})$ stored in the replay buffer is not only with the original goal $g_t$ used for the episode but also with a subset of other goals $(g_{t_1}',g_{t_2}',\ldots,g_{t_k}')$, and the additional transitions with new goals are relabeled as $(s_{t},a_{t},g_{t_1}',s_{t+1},r_{t_1}')$, $(s_{t},a_{t},g_{t_2}',s_{t+1},r_{t_2}'), \ldots,$$(s_{t},a_{t},g_{t_k}',s_{t+1},r_{t_k}')$ with recalculated reward $r_{t_1}',r_{t_2}',\ldots,r_{t_k}'$. Here hyperparameter $k$ controls the ratio of HER data to data coming from regular experience replay in the replay buffer. In [10], four strategies are supplied to sample additional goals to replay, including $\emph{final}$, $\emph{future}$, $\emph{episode}$ and $\emph{random}$. Empirical results show that in all cases, $\emph{future}$ with $k$ equal 4 or 8 has the best performance. HER may be seen as a form of implicit curriculum to achieve goals from simple to more difficult ones.

\section{METHOD}

In this section, we present the detailed architecture of Augmented Curiosity-Driven Experience Replay (ACDER), in which we use two methods to improve the exploration performance: goal-oriented curiosity-driven exploration and dynamic initial start state selection.

\subsection{Goal-Oriented Curiosity-Driven Exploration}
First, we utilize S-ICM [40], a simplified intrinsic curiosity module, to generate curiosity rewards. In S-ICM, it uses a forward dynamics model $\emph{F}:\mathcal{S}\times\mathcal{A}\rightarrow\mathcal{S}$ parameterized by $\theta^F$ to predict next state given current state and action and defines the state prediction error as an additional intrinsic curiosity reward. When facing unfamiliar states, the agent will receive high intrinsic rewards, which facilitates it to seek novel states and thus increases the chance of stumbling into the goal states. Different from combining curiosity-driven exploration with on-policy reinforcement learning algorithm [16], which results in the agent forgetting the visited paths and exploring those states repeatedly, S-ICM enables sampling past experiences from the replay buffer for training. In this way, it can ensure exploration stability and avoid inefficient exploration behavior, especially in large-scale environments with sparse rewards [41].

For each transition $(s_t,a_t,g^h,s_{t+1},r^e_t)$, the intrinsic curiosity reward $r^i_t$ is computed as the squared error between the predicted next state $F(s_t,a_t|\theta^F)$ and the actual next state $s_{t+1}$, which can be written as:
$$
r^i_t=clip(\frac{1}{2}{\left\|F(s_t,a_t|\theta^F)-s_{t+1}\right\|}^2_2,0,\eta) \eqno{(4)}
$$
$0<\eta<1$ is a scaling factor that limits the magnitude range of curiosity reward. $\emph{clip}$ function is used to keep novelty gap between states to encourage the agent to pursue more novel states purposefully. 

\begin{figure}[thpb]
\centering
\subfigure{\includegraphics[width=8cm]{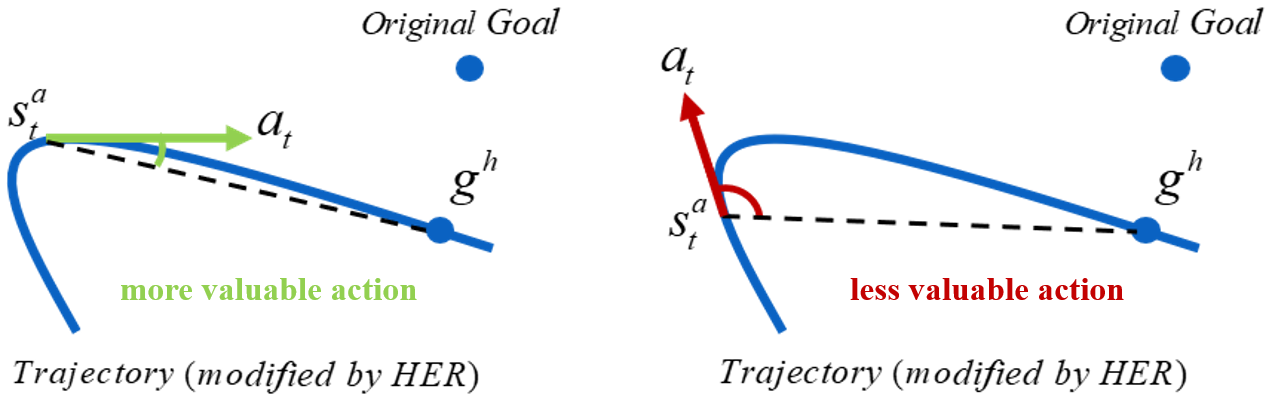}}
\caption{The illustration of goal-oriented factor.}
\label{figurelabel}
\end{figure}
Second, to avoid blindly pursuing novel states, we aim to encourage more task-relevant exploration, which helps to meet more novel states relevant to specific goals. For transitions after HER modification, we introduce a goal-oriented factor that is determined by the angle between action vector $\overrightarrow{a_t}$ and the vector from current robotic agent position to new additional goal position $\overrightarrow{s^{a}_t g^h}$ to form new extrinsic rewards $r^{e*}_t$, as follows:
$$
r^{e*}_t= \lambda_r \times r(s_t,a_t,g^h)\eqno{(5)}
$$
where
$$
\lambda_r = \frac{1}{2\pi}\arccos<\overrightarrow{s^{a}_t g^h},\overrightarrow{a_t}> + 1\eqno{(6)}
$$
$\lambda_r$ is the goal-oriented factor that linearly increases as the angle between the two vectors increases (illustrated in Fig. 1) and $r(s_t,a_t,g^h)\in\{-1,0\}$ is the task reward after HER modification. In this way, the actions with larger rewards $r^{e*}_t$ become more valuable to the new additional goals and would improve the learning efficiency for the generated goals $g^h$, and then further accelerate the progress of achieving the original goals. Combining intrinsic curiosity rewards and the task-relevant new extrinsic task rewards, we can encourage the agent to seek the states that are novel and surprising and also relevant to goals for exploration.

At each update step, the goal-oriented curiosity model and RL policy are trained simultaneously using the same transitions, which can make the model and policy concurrently converge as the agent gains more knowledge about its environment. The overall optimization objective is to maximize the expected sum of joint reward composed of intrinsic curiosity reward and new goal-oriented extrinsic task reward and minimize loss function $L_F$ of the forward dynamics neural network in S-ICM, which can be written as:
$$
\mathop{\min}\limits_{\theta^P\theta^F}\left[-\mathbb{E}_{\pi(s_t|\theta^P)}[\sum_{t}r^{e*}_t+r^i_t]+L_F\right] \eqno{(7)}
$$
where
$$
L_F=\frac{1}{2}{\left\|F(s_t,a_t|\theta^F)-s_{t+1}\right\|}^2_2 \eqno{(8)}
$$

\subsection{Dynamic Initial State Selection}
For robotic manipulation tasks, sometimes the agent is hard to visit the whole state space if it always starts from a fixed state in limited step horizons. To further improve exploration efficiency, we enable the agent to start from those novel or rare states experienced before rather than always a fixed initial state. As the agent has already gained ability to reach some states, it is unnecessary to always start from the scratch [42]. Increasing visited state diversity can enhance agent exploration efficiency and learning performance. Obviously, it is more likely to find novel states within the range of novel states. By starting from unfamiliar states that the agent has already encountered, we could create an automatic exploratory curriculum that helps the agent gradually explore state space and even visit those beyond the originally reachable states.

The initial states selected to restart from in the next episode should have two properties: (i) the states are novel compared with those stored currently in replay buffer; (ii) the agent has the capability to reach the states under current policy. To accomplish these, we first utilize S-ICM to generate intrinsic curiosity reward for each sampled transition. Then we extract transitions that stem from successful episodes in replay buffer from these training samples. In the extracted transitions, we further select the state which has maximal intrinsic reward as an eligible state to restart from in the next episode.

In order to avoid policy overfitting and collapse, we maintain a balance between selecting novel states and the original fixed initial state, which can ensure to both further diversify the initial states and prevent performance degradation in the later learning process. To achieve this objective, we utilize a hyperparameter $\alpha$ to determine the ratio of selecting novel states to restart for the next episode. It is tuned dynamically according to the test success rate throughout training, which is defined as:
$$
\alpha = \alpha_0-0.5\times(10e)^{success\_rate-1} \eqno{(9)}
$$
where $\alpha_0$ is the initial ratio. In the early stage of learning, generating episodes from novel states can help explore the environment faster compared with random exploration. As the policy updates, the ratio $\alpha$ will be set to prefer more experiences from the original fixed initial states for optimizing the original performance metric. By dynamically adjusting the rate to select initial states to generate episodes, we aim to increase initial state diversity for accelerating coverage of the state space and improving task performance.

\subsection{Architecture Details}
\begin{figure}[thpb]
\centering
\subfigure{\includegraphics[width=7cm]{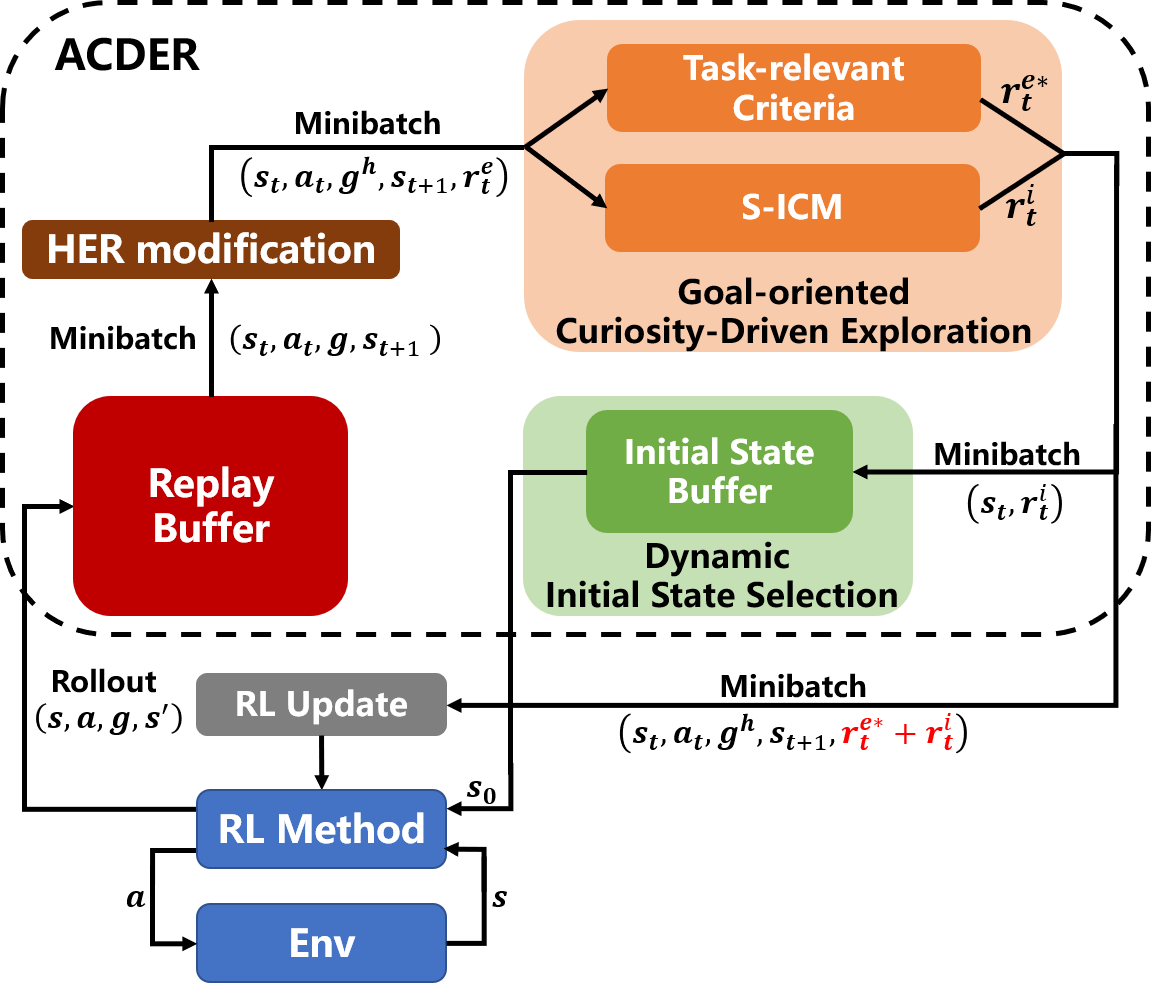}}
\caption{Schematic view of ACDER.}
\label{figurelabel}
\end{figure}

A schematic view of ACDER is given in Fig. 2. At each update, rollouts produced by agent's interacting with the environment are stored in the replay buffer. After HER modification, the training samples are passed into the S-ICM to obtain respective intrinsic curiosity rewards and form the new goal-oriented extrinsic rewards based on task-relevant criteria. Combining intrinsic rewards and the new extrinsic rewards, the agent is encouraged to seek more task-relevant and novel states for fast convergence. Moreover, we utilize a dynamic initial state selection module to start the agent from novel states with high intrinsic rewards to further increase state diversity. 

ACDER can be potentially used with any off-policy RL algorithm. In our experiments, we use DDPG for policy learning, and our method can indeed improve exploration efficiency and provide an order of magnitude of speedup over RL method.

\section{EXPERIMENTS AND RESULTS}

In this section, we first introduce the robotic simulation environments we use for the experiments. Second, we compare the performance of our method combined with DDPG, with four existing RL algorithms. Third, we check if our approach improves sample efficiency in robotic manipulation tasks. Fourth, ablation environments show the importance of each component on learning performance. Finally, we show the results of the experiments on the physical robot. 

\subsection{Simulation Environments}

The environment we used throughout our experiments is a robotic simulation provided by OpenAI Gym [8, 43], which introduces a suite of challenging continuous control tasks based on current existing robotic hardware. In all experiments, we use a robotic agent that is a 7-DOF Fetch robotic arm with a two-fingered parallel gripper. The robot is simulated using the MuJoCo physics engine [44]. We evaluate our method in four challenging tasks, including Reach, Push, Pick$\emph{\&}$Place, and Multi-Step Push, see Fig. 3.

In Fetch tasks, states include the positions and linear velocities of the gripper and fingers. If an object is present, it also includes the positions, rotations, and angular velocities of object, as well as its position and linear velocities relative to the gripper. Action is 4-dimensional: 3 dimensions output gripper incremental movement and the last dimension controls the opening and closing of fingers. Goal is the desired position of end-effector or goal-object (if any). Rewards are sparse and binary: the agent receives a reward of 0 if the goal is reached (within a distance of 5 cm) and -1 otherwise.

\begin{figure*}[!t]
\centering
\subfigure{\includegraphics[width=3cm]{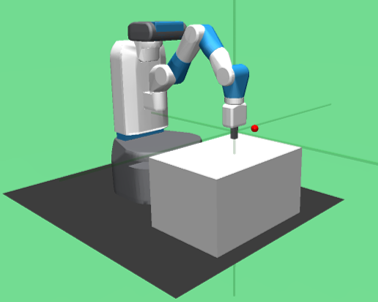}}
\hspace{1cm}
\subfigure{\includegraphics[width=3cm]{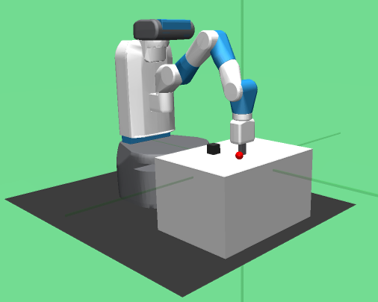}}
\hspace{1cm}
\subfigure{\includegraphics[width=3cm]{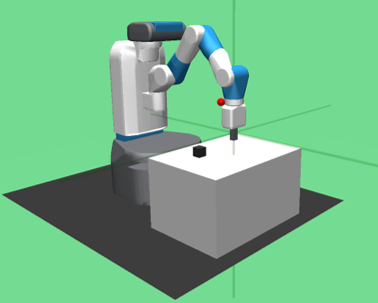}}
\hspace{1cm}
\subfigure{\includegraphics[width=3cm]{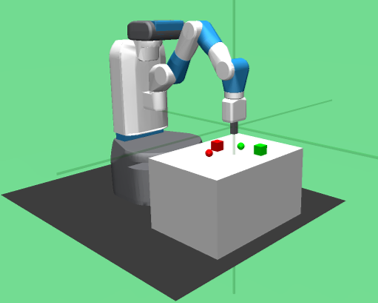}}
\caption{ Robotic simulation environment: FetchReach, FetchPush, FetchPickAndPlace and Multi-Step FetchPush.}
\end{figure*}

\begin{figure*}[!t]
\centering
\subfigure{\includegraphics[width=4.1cm]{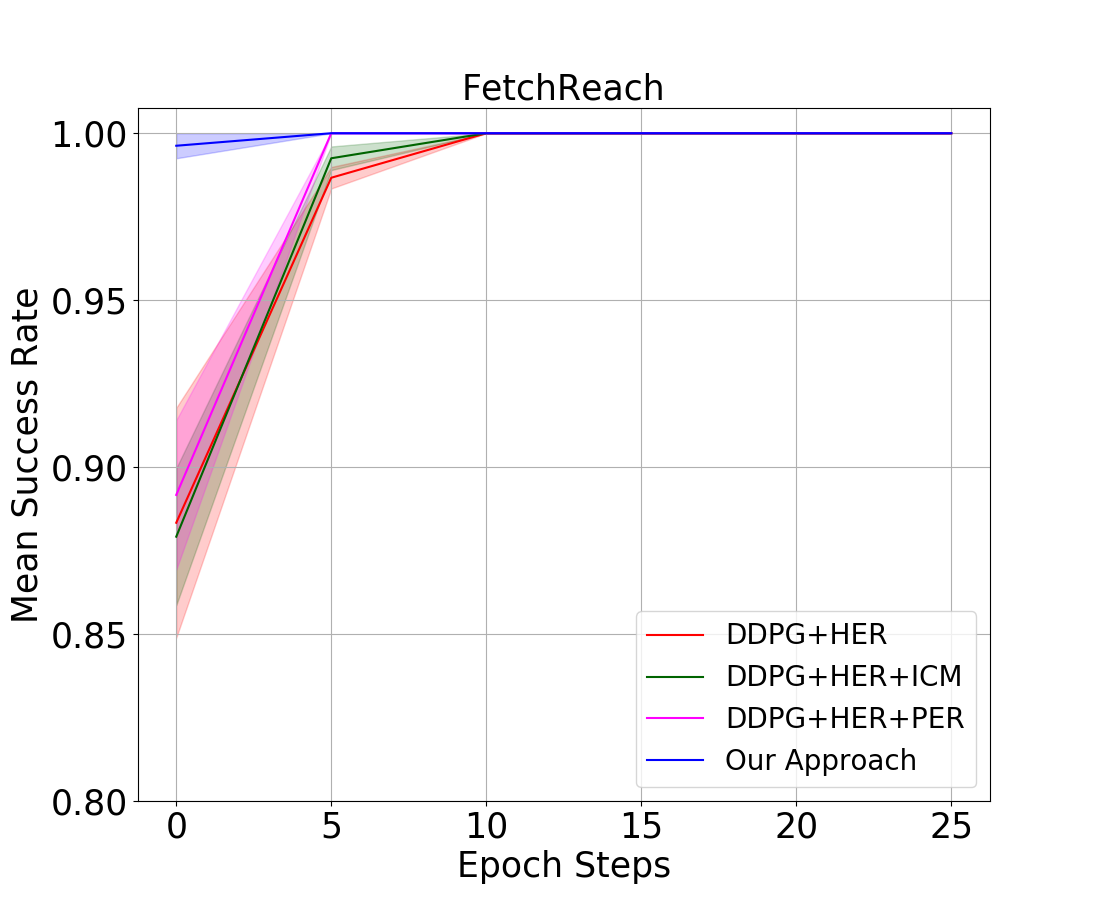}}
\subfigure{\includegraphics[width=4.1cm]{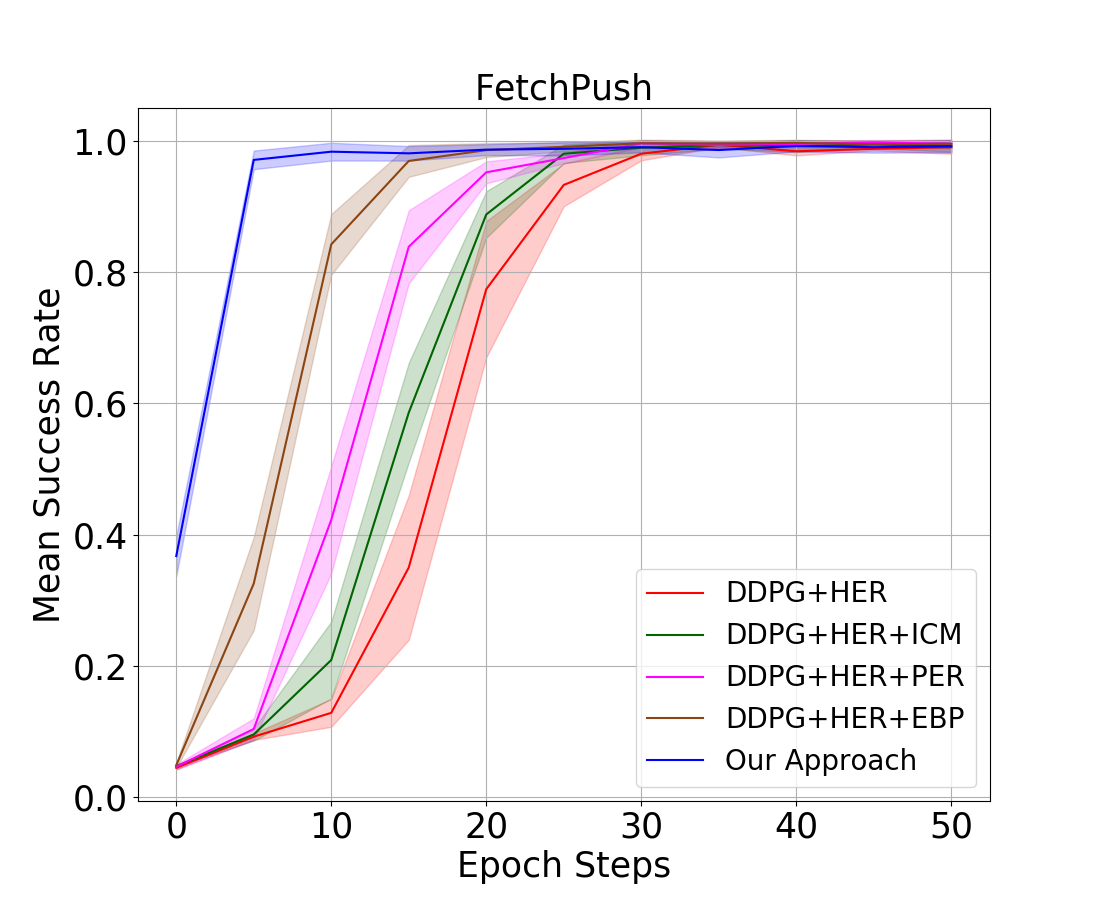}}
\subfigure{\includegraphics[width=4.1cm]{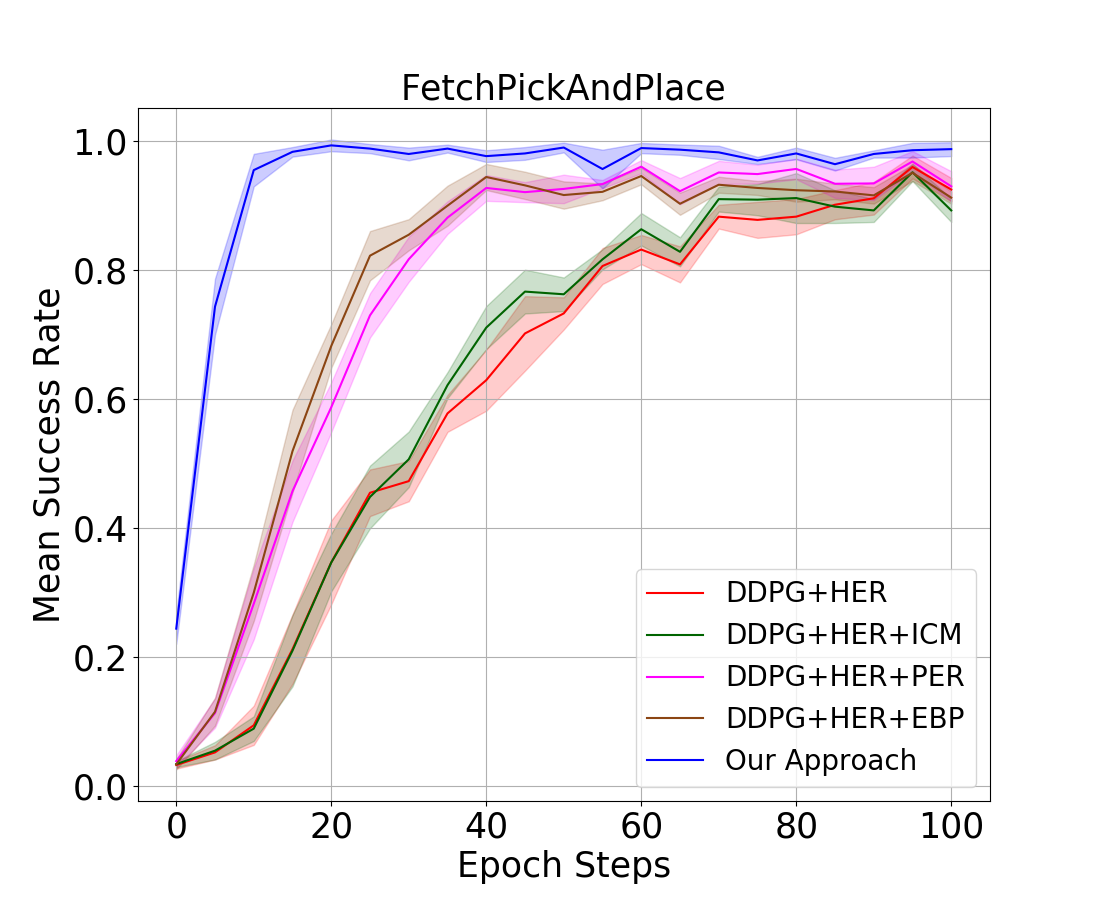}}
\subfigure{\includegraphics[width=4.1cm]{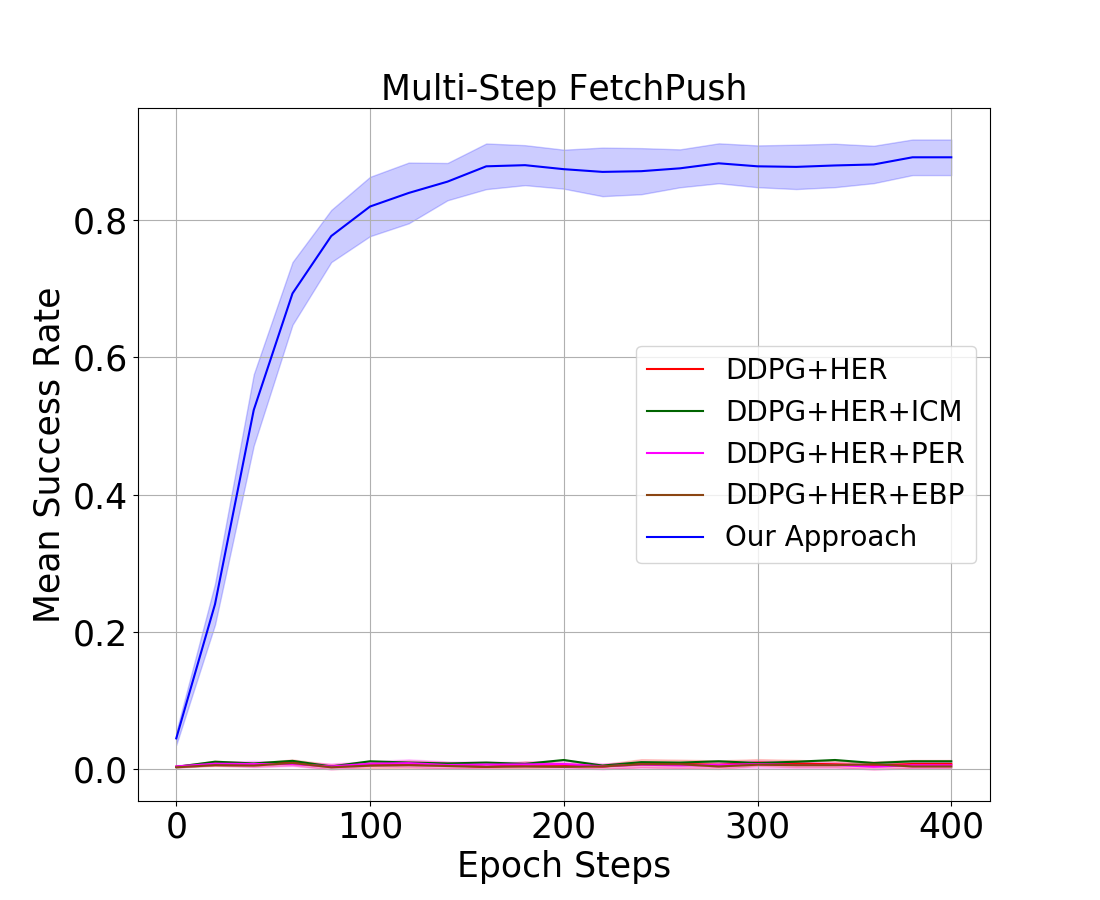}}
\caption{ Comparison on mean test success rate of DDPG+HER+ACDER and other existing methods in all four environments. }
\end{figure*}

\subsection{Performance}
In order to test the performance, we evaluate the difference between our method ACDER with DDPG and vanilla DDPG+HER (OpenAI Baselines) on all four robotic tasks. Moreover, we compare against DDPG+HER with PER [34] which prioritizes the transitions with high TD-error for hindsight transformation and DDPG+HER with EBP [37] which prioritizes trajectories with more physical work-energy done and then sampled transitions for hindsight transformation. These two methods are both existing improvements of experience replay which have been proven to improve the learning performance and sample efficiency. We also compare against DDPG+HER with ICM [16], which encourages the agent to seek novel states for exploration by defining state prediction error as intrinsic rewards. In all methods, HER utilizes $\emph{future}$ strategy with $k=4$ for replacing original goals with achieved goals that observed later from the same episodes.

We compare the mean test success rate of these five methods. Because of no object, we only compare four methods in the FetchReach environment, without DDPG+HER+EBP. Each experiment is carried out across 5 random seeds, and the shadow area represents one standard deviation. In all experiments, we use 4 CPUs and train the agent for 25 epochs (each epoch includes $4\times100$ episodes) in the FetchReach environment, 50 epochs in the FetchPush environment, 100 epochs in the FetchPickAndPlace environment and 400 epochs in the Multi-Step Push environment. Throughout the experiments, $\eta=0.05$ is used for Eq(4) in the Reach task, while $\eta=0.8$ in the other three remaining tasks. The initial ratio $\alpha_0$ in Eq(9) is set to 0.8 in three basic environments and 0.7 in the multi-step task. The forward dynamics network in S-ICM is a three-layer fully connected framework. The input layer is constructed by concatenating $(s_t,g)$ with $a_t$. The hidden layer includes 256 units, and the number of the final output units is the total dimensions of state and goal of the task. The parameters of actor and critic network in DDPG are all the same as those in DDPG+HER (OpenAI Baselines). We use a replay buffer with a size of $10^6$ and train the policy with minibatch sizes of 256. In this way, we hope the forward dynamics model fits the familiar training states but keeps insensitive to unseen states, which can help to determine state novelty. The learning curve with respect to training epochs is shown in Fig. 4. We can see clearly that our method, DDPG+ACDER offers faster convergence and better improvement of final success rate than the other four methods in all four robotic tasks without requiring any increase in the number of episodes. Moreover, our approach enables multi-step robotic manipulation task learning, while the other four methods can't.

\begin{table}[h]
\caption{Number of episodes needed for a certain test success rate in all three environments}
\label{table_example}
\begin{center}
\linespread{1.5}\selectfont
\begin{tabular}{|c||c||c||c|}
\hline
 & \makecell[c]{Reach\\(100$\%$)} & \makecell[c]{Push\\(95$\%$)} & \makecell[c]{Pick\emph{\&}Place\\(95$\%$)} \\
\hline
DDPG+HER & 2,240 & 11,600 & 33,600\\
\hline
DDPG+HER+ICM & 2,400 & 9,800 & 29,800\\
\hline
DDPG+HER+PER & 1,440 & 7,800 & 17,400\\
\hline
DDPG+HER+EBP & $-$ & 7,200 & 17,200\\
\hline
\textbf{DDPG+ACDER} & \textbf{400} & \textbf{2,200} & \textbf{5,200}\\
\hline
\end{tabular}
\end{center}
\end{table} 

\begin{figure*}[!t]
\centering
\subfigure{\includegraphics[width=4.1cm]{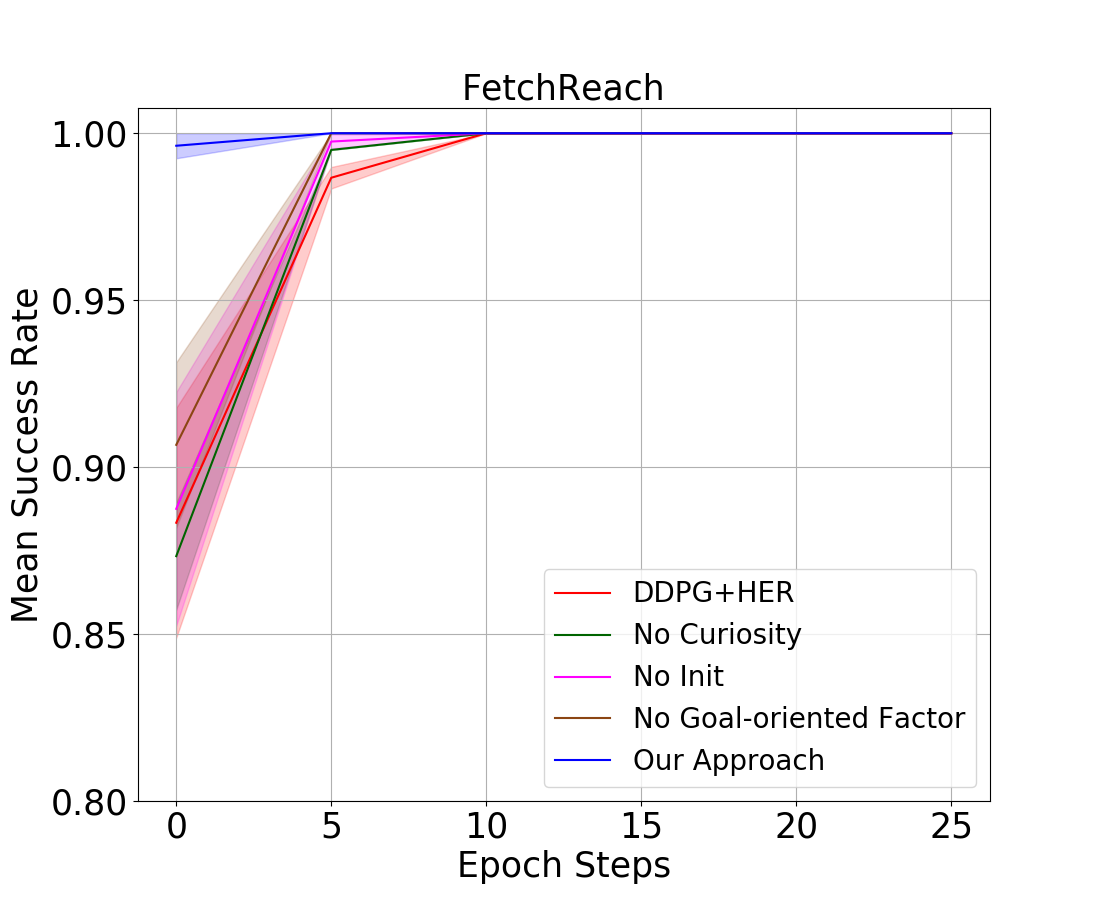}}
\subfigure{\includegraphics[width=4.1cm]{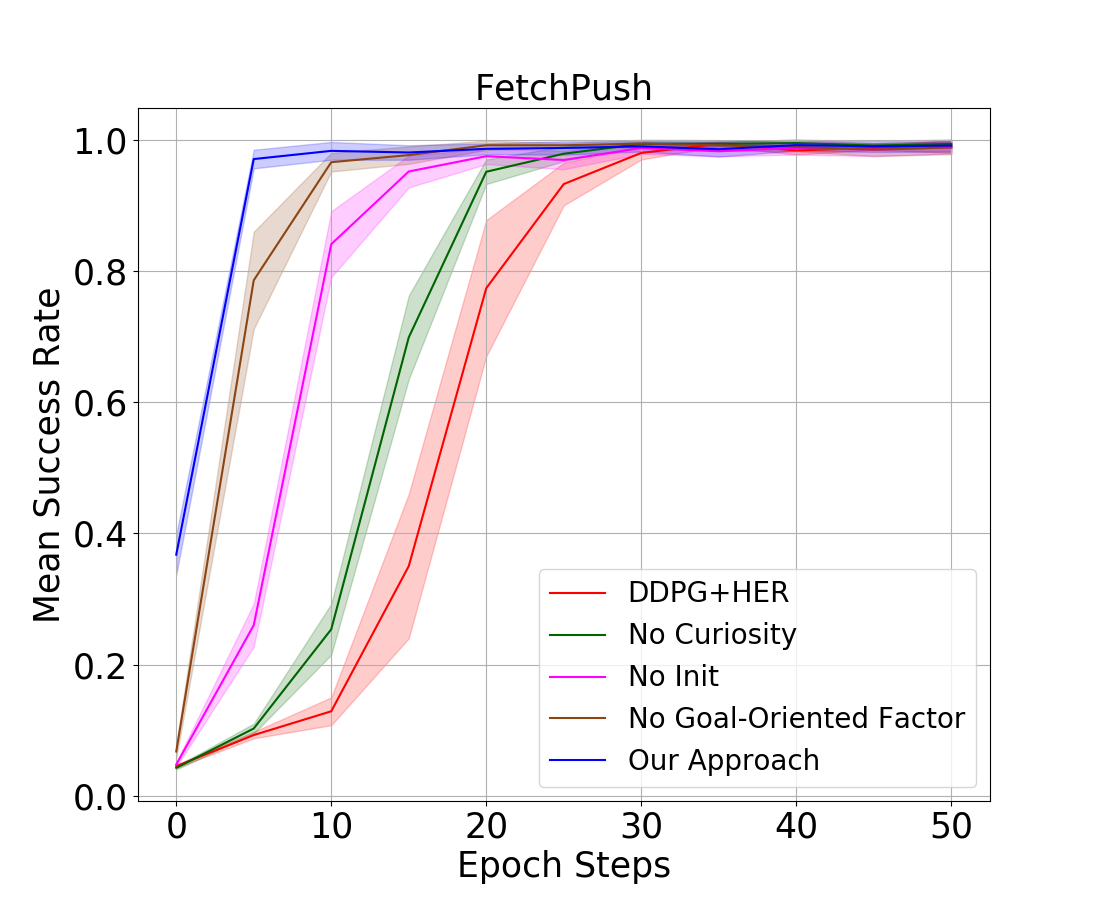}}
\subfigure{\includegraphics[width=4.1cm]{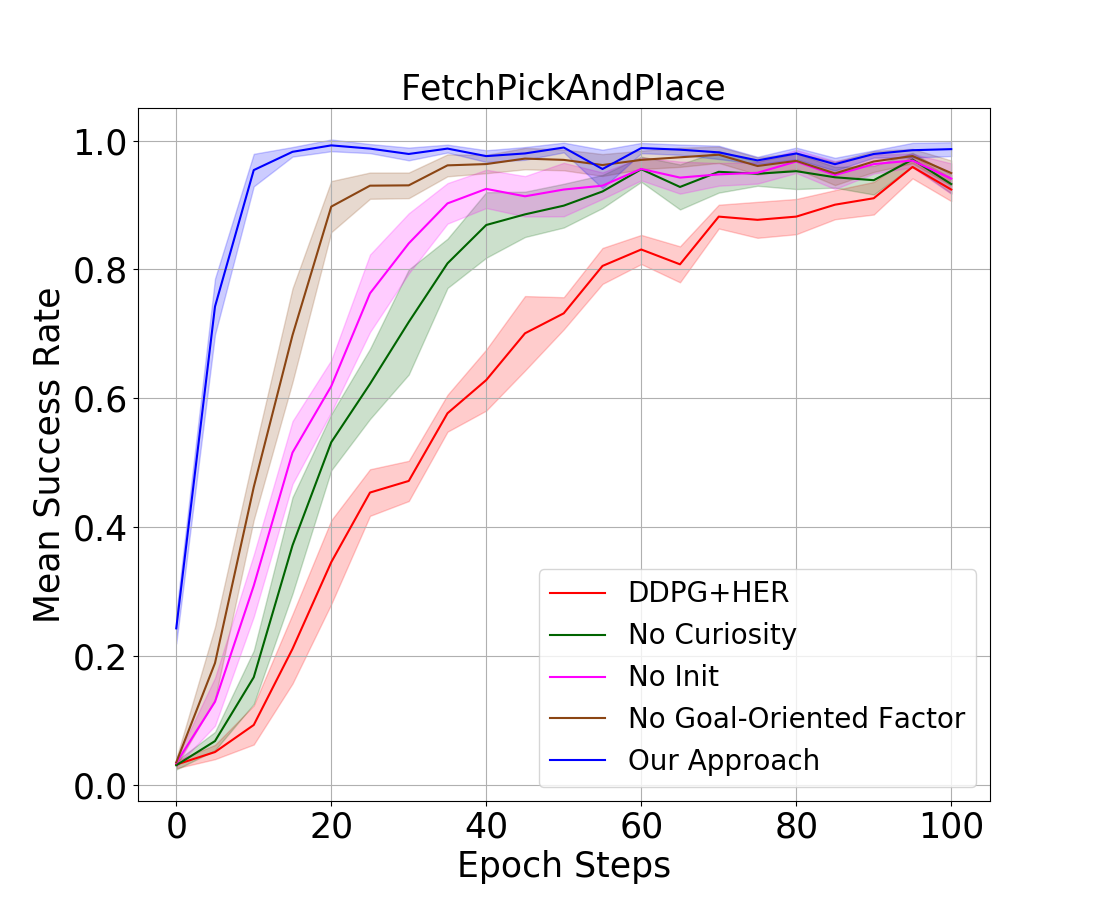}}
\caption{ Ablation results on first three basic robotic environments.}
\end{figure*}

\subsection{Sample Efficiency}
To verify if our method improves sample efficiency, we compare the number of episodes needed in the replay buffer when achieving the same mean test success rate using these five methods. From Table I, we can observe that in three basic robotic tasks, DDPG+ACDER always needs the minimal training samples to reach the same success rate compared with the other four methods. Especially in relatively harder Pick\emph{\&}Place task, for the same 95$\%$ test success rate, vanilla DDPG+HER needs 33,600 episodes for training, while DDPG+ACDER only needs 5,200 episodes, which the sample size has been decreased by more than five times. Meanwhile, compared with two existing improvements of DDPG+HER, our method also nearly triples the sample-efficiency. In conclusion, DDPG+ACDER can improve sample efficiency remarkably by an average factor of five over vanilla DDPG+HER.

\subsection{Ablation Experiments}
We perform the ablation experiments to measure the importance of each component of our methods on learning performance. From Fig. 5 we can see clearly that without curiosity-driven exploration, the method is significantly worse in each task, but still learns faster than vanilla DDPG+HER. Similarly, without dynamic initial state selection, the method's learning performance also suffers but achieves much better than vanilla DDPG+HER. Moreover, without goal-oriented factor, pursuing novel states blindly will hinder the agent's learning.

The goal-oriented curiosity-driven exploration is an effective method to accelerate exploration and achieve optimal learning performance by introducing intrinsic rewards to encourage the agent to seek task-relevant and novel states more purposefully. Dynamic initial state selection enables the agent to gradually explores the environment in a directed way according to the agent's knowledge about it. Therefore, combining goal-oriented curiosity-driven exploration and initial state selection, our method is expected to maximize exploration efficiency and improve learning performance.

\subsection{Deployment on a Physical Robot}
For easy deployment on a physical robot, we retrain the policies for three basic robotic tasks (Reach, Push, Pick\emph{\&}Place) with simplified state inputs in the simulation environments with the 7-DOF Fetch robot. The states only include the positions of gripper and objects, the relative positions of objects to gripper and finger states. And then we deploy the trained policies on a real-world UR5 robotic arm respectively. Since the output of policy is the incremental movement of the end-effector of the robotic arm, it is feasible to transfer the policies directly without consideration of the robot configuration. The object and goal positions are obtained by color recognition using raw RGB and depth camera images [45]. Fig. 6 shows the process of policy deployed on real-world UR5 robotic arm. The red cube represents the target position in reach task, the green cylinder is pushed upon the red target in push task, and the orange cube is placed on the yellow cube in pick\emph{\&}place task. We test the success rate with 5 cm tolerance error in each task for 20 trials. The policies succeed in 19 out of 20 trials in reach task, 16 out of 20 trials in push task, while 17 out of 20 in pick\emph{\&}place task, which shows the consistent performance between simulation environment and real-world scenarios.

\begin{figure}[thpb]
\centering
\subfigure{\includegraphics[width=1.6cm]{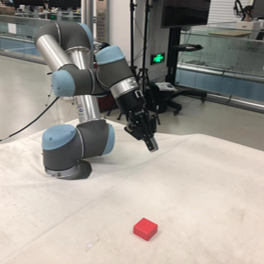}}
\subfigure{\includegraphics[width=1.6cm]{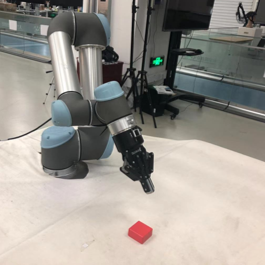}}
\subfigure{\includegraphics[width=1.6cm]{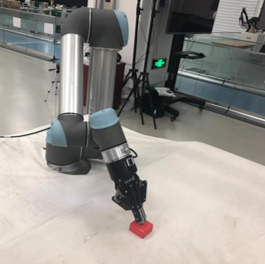}}
\subfigure{\includegraphics[width=1.6cm]{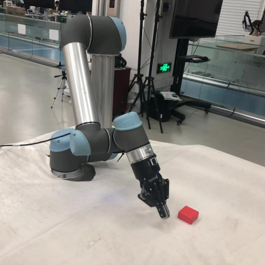}}
\subfigure{\includegraphics[width=1.6cm]{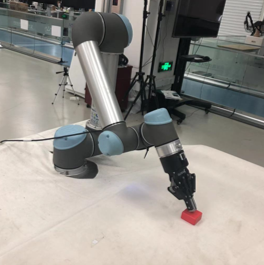}}
\centerline{(a)}

\subfigure{\includegraphics[width=1.6cm]{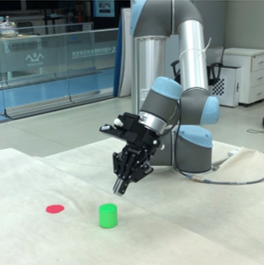}}
\subfigure{\includegraphics[width=1.6cm]{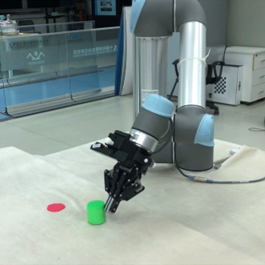}}
\subfigure{\includegraphics[width=1.6cm]{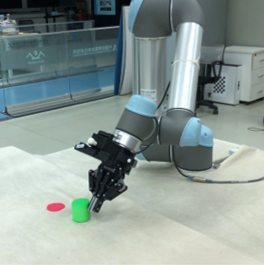}}
\subfigure{\includegraphics[width=1.6cm]{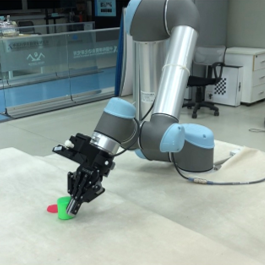}}
\subfigure{\includegraphics[width=1.6cm]{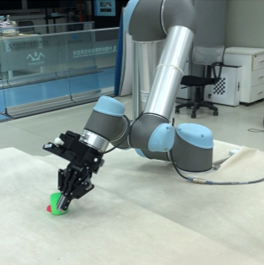}}
\centerline{(b)}

\subfigure{\includegraphics[width=1.6cm]{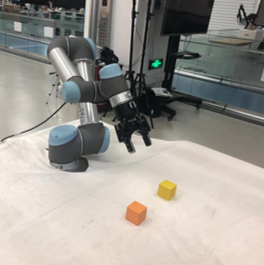}}
\subfigure{\includegraphics[width=1.6cm]{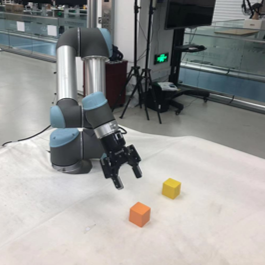}}
\subfigure{\includegraphics[width=1.6cm]{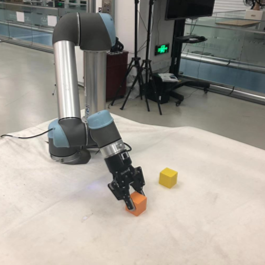}}
\subfigure{\includegraphics[width=1.6cm]{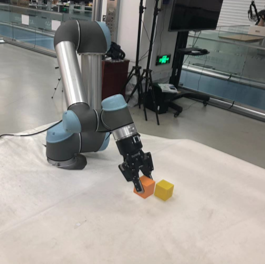}}
\subfigure{\includegraphics[width=1.6cm]{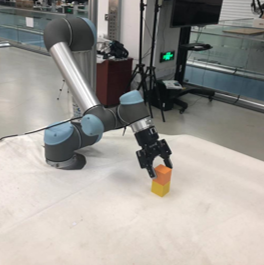}}
\centerline{(c)}

\caption{Frames of learned policy on real-world UR5 robotic arm for (a) reach task, (b) push task and (c) pick\emph{\&}place task.}
\label{figurelabel}
\end{figure}

\section{CONCLUSIONS} 

We propose an effective method called Augmented Curiosity-Driven Experience Replay (ACDER) which introduces a goal-oriented curiosity-driven exploration to encourage the agent to pursue task-relevant and novel states more efficiently and purposefully, and learn policy through an automatic exploratory curriculum by dynamic initial start states selection. ACDER shows promising experimental results in all three basic challenging robotic manipulation tasks and improves sample efficiency by a factor of five over vanilla DDPG+HER. And our method also has the capability to achieve multi-step task learning. Moreover, we show that the policies trained in simulation for reach, push and pick$\emph{\&}$place tasks perform well on the physical robot without any additional finetuning.

In future works, we would employ our method in higher dimensional continuous environments. A promising research is to utilize this idea to solve exploration problems in more complicated robotic manipulation tasks with longer horizons.







\newpage
\begin{thebibliography}{99}

\bibitem{c1} V. Mnih, K. Kavukcuoglu, D. Sliver, et al. Human-level control through deep reinforcement learning. Nature, vol. 518, no. 7540, pp. 529-533, 2015.
\bibitem{c2} J. Schulman, S. Levine, P. Moritz, et al. Trust region policy optimization. International Conference on Machine Learning, pp. 1889-1897, 2015.
\bibitem{c3} J. Schulman, F. Wolski, P. Dhariwal, et al. Proximal policy optimization algorithms. arXiv preprint arXiv:1707.06347, 2017.
\bibitem{c4} A. Y. Ng, A. Coates, M. Diel, et al. Autonomous inverted helicopter flight via reinforcement learning. Experimental Robotics IX, pp. 363-372, Springer, 2006.
\bibitem{c5} Y. Chebotar, M. Kalakrishnan, A. Yahya, et al. Path integral guided policy search. IEEE International Conference on Robotics and Automation, pp. 3381-3388, 2017.
\bibitem{c6} S. Levine, P. Pastor, A. Krizhevsky, et al. Learning hand-eye coordination for robotic grasping with deep learning and large-scale data collection. The International Journal of Robotics Research, vol. 37, no. 4-5, pp. 421-436, 2018.
\bibitem{c7} D. Kalashnikov, A. Irpan, P. Pastor, et al. Qt-opt: Scalable deep reinforcement learning for vision-based robotic manipulation. arXiv preprint arXiv:1806.10293, 2018.
\bibitem{c8} M. Plappert, M. Andrychowicz, A. Ray, et al. Multi-goal reinforcement learning: Challenging robotics environments and request for research. arXiv preprint arXiv:1802.09464, 2018.
\bibitem{c9} I. Popov, N. Heess, T. Lillicrap, et al. Data-efficient deep reinforcement learning for dexterous manipulation. arXiv preprint arXiv:1704.03073, 2017.
\bibitem{c10} M. Andrychowicz, F. Wolski, A. Ray, et al. Hindsight experience replay. Advances in Neural Information Processing Systems, pp. 5048-5058, 2017.
\bibitem{c11} R. S. Sutton and A. G. Barto. Reinforcement learning: An introduction. MIT Press, 1998.
\bibitem{c12} Y. Li. Deep reinforcement learning. arXiv preprint arXiv:1810.06339, 2018.
\bibitem{c13} M. Riedmiller, R. Hafner, T. Lampe, et al. Learning by playing-solving sparse reward tasks from scratch. arXiv preprint arXiv:1802.10567, 2018.
\bibitem{c14} P. Oudeyer. Computational theories of curiosity-driven learning. arXiv preprint arXiv:1802.10546, 2018.
\bibitem{c15} M. Gregor, J. Spalek. Curiosity-driven exploration in reinforcement learning. Physiology \emph{\&} Behavior, vol. 6, no. 1, pp. 435-440, 2014.
\bibitem{c16} D. Pathak, P. Agrawal, A. A. Efros, et al. Curiosity-driven exploration by self-supervised prediction. International Conference on Machine Learning, pp. 2778-2787, 2017.
\bibitem{c17} N. Savinov, A. Raichuk, R. Marinier, et al. Episodic curiosity through reachability. arXiv preprint arXiv:1810.02274, 2018.
\bibitem{c18} T. P. Lillicrap, J. J. Hunt, A. Pritzel, et al. Continuous control with deep reinforcement learning. arXiv preprint arXiv:1509.02971, 2015.
\bibitem{c19} T. L. Lai, H. Robbins. Asymptotically efficient adaptive allocation rules. Advances in Applied Mathematics, vol. 6, no. 1, pp. 4-22, 1985.
\bibitem{c20} A. L. Strehl, M. L. Littman. An analysis of model-based interval estimation for Markov decision processes. Journal of Computer and System Sciences, vol. 74, no. 8, pp. 1309-1331, 2008.
\bibitem{c21} Z. Xu, X. Chen, L. Cao, et al. A study of count-based exploration and bonus for reinforcement learning. IEEE International Conference on Cloud Computing and Big Data Analysis, pp. 425-429, 2017.
\bibitem{c22} M. G. Bellemare, S. Srinivasan, G. Ostrovski, et al. Unifying count-based exploration and intrinsic motivation. Advances in Neural Information Processing Systems, pp. 1471-1479, 2016.
\bibitem{c23} G. Ostrovski, M. G. Bellemare, A. Oord, et al. Count-based exploration with neural density models. International Conference on Machine Learning, pp. 2721-2730, 2017.
\bibitem{c24} H. Tang, R. Houthooft, D. Foote, et al. $\#$Exploration: A study of count-based exploration for deep reinforcement learning. Advances in Neural Information Processing Systems, pp. 2753-2762, 2017.
\bibitem{c25} J. Fu, J. Co-Reyes, S. Levine. EX$^2$: Exploration with exemplar models for deep reinforcement learning. Advances in Neural Information Processing Systems, pp. 2577-2587, 2017.
\bibitem{c26} C. Stanton, J. Clune. Deep curiosity search: Intra-life exploration can improve performance on challenging deep reinforcement learning problems. arXiv preprint arXiv:1806.00553, 2018.
\bibitem{c27} S. H. Huang, M. Zambelli, J. Kay, et al. Learning gentle object manipulation with curiosity-driven deep reinforcement learning. arXiv preprint arXiv:1903.08542, 2019.
\bibitem{c28} Y. Burda, H. Edwards, D. Pathak, et al. Large-scale study of curiosity-driven learning. arXiv preprint arXiv:1808.04355, 2018.
\bibitem{c29} Y. Burda, H. Edwards, A. Storkey, et al. Exploration by random network distillation. arXiv preprint arXiv:1810.12894, 2018.
\bibitem{c30} L. J. Lin. Self-improving reactive agents based on reinforcement learning, planning and teaching. Machine learning, vol. 8, no. 3-4, pp. 293-321, 1992.
\bibitem{c31} S. Gu, T. Lillicrap,I. Sutskever, et al. Continuous deep q-learning with model-based acceleration. International Conference on Machine Learning, pp. 2829-2838, 2016.
\bibitem{c32} Z. Wang, V. Bapst, N. Heess, et al. Sample efficient actor-critic with experience replay. arXiv preprint arXiv:1611.01224, 2016.
\bibitem{c33} S. Lanka, T. Wu. ARCHER: Aggressive rewards to counter bias in hindsight experience replay. arXiv preprint arXiv:1809.02070, 2018.
\bibitem{c34} T. Schaul, J. Quan, I. Antonoglou, et al. Prioritized experience replay. arXiv preprint arXiv:1511.05952, 2015.
\bibitem{c35} Y. Hou, L. Liu, Q. Wei, et al. A novel ddpg method with prioritized experience replay. IEEE International Conference on Systems, Man, and Cybernetics, pp. 316-321, 2017. 
\bibitem{c36} M. Fang, C. Zhou, B. Shi, et al. DHER: Hindsight experience replay for dynamic goals. International Conference on Robot Learning, 2019.
\bibitem{c37} R. Zhao, V. Tresp. Energy-based hindsight experience prioritization. arXiv preprint arXiv:1810.01363, 2018.
\bibitem{c38} R. Zhao, V. Tresp. Curiosity-driven experience prioritization via density estimation. arXiv preprint arXiv:1902.08039, 2019.
\bibitem{c39} T. Schaul, D. Horgan, K. Gregor, et al. Universal value function approximators. International Conference on Machine Learning, pp. 1312-1320, 2015.
\bibitem{c40} B. Li, T. Lu, J. Li, et al. Curiosity-driven exploration for off-policy reinforcement learning methods. IEEE International Conference on Robotics and Biomimetics, pp. 1109-1114, 2019.
\bibitem{c41} H K. Yang, P H. Chiang, K W. Ho, et al. Never forget: Balancing exploration and exploitation via learning optical flow. arXiv preprint arXiv:1901.08486, 2019.
\bibitem{c42} A. Tavakoli, V. Levdik, R. Islam, et al. Prioritizing starting states for reinforcement learning. arXiv preprint arXiv:1811.11298, 2018.
\bibitem{c43} G. Brockman, V. Cheung, L. Pettersson, et al. Openai gym. arXiv preprint arXiv:1606.01540, 2016.
\bibitem{c44} E. Todorov, T. Erez, Y. Tassa. Mujoco: A physics engine for model-based control. IEEE/RSJ International Conference on Intelligent Robots and Systems, pp. 5026-5033, 2012.
\bibitem{c45} B. Li, T. Lu, X. Li, et al. An automatic robot skills learning system from robot's real-world demonstrations. Chinese Control And Decision Conference, pp. 5138-5142, 2019.
\end{thebibliography}
\end{document}